\documentclass{ametsocV5}

\usepackage{amsfonts,amssymb,bm}
\usepackage{mathptmx}%{times}
\usepackage{newtxtext}
\usepackage{newtxmath}
\usepackage{multicol}
\usepackage{dsfont}
\usepackage{bbm}
\usepackage{subcaption}
\usepackage{siunitx}
\newcommand{\ind}{\mathbbm{1}}

\usepackage{tikz}
\usetikzlibrary{positioning}

\usepackage{multirow}
\usepackage{newfloat}
\newcommand{\sjoerd}[1]{{\color{black}{#1}}}
\newcommand{\sjoerdr}[1]{{\color{black}{#1}}}
\newcommand{\simon}[1]{{\color{black}{#1}}}
\newcommand{\maurice}[1]{{\color{black}{#1}}}
\newcommand{\mauricer}[1]{{\color{black}{#1}}}
\newcommand{\kiri}[1]{{\color{black}{#1}}}

\title{Statistical post-processing of wind speed forecasts using convolutional neural networks}
\authors{Simon Veldkamp\thanks{Royal Netherlands Meteorological Institute (KNMI), the Netherlands}, Kiri\sjoerd{en} Whan$^*$, Sjoerd Dirksen\thanks{Mathematical Institute, Utrecht University, the Netherlands} and Maurice Schmeits$^*$\correspondingauthor{Maurice Schmeits, maurice.schmeits@knmi.nl}}

\authors{Simon Veldkamp}%\thanks{Royal Netherlands Meteorological Institute (KNMI), the Netherlands}}
\affiliation{Royal Netherlands Meteorological Institute (KNMI), De Bilt, the Netherlands, and Mathematical Institute, Utrecht University, the Netherlands}
\extraauthor{Kirien Whan}
\extraaffil{Royal Netherlands Meteorological Institute (KNMI), De Bilt, the Netherlands}
\extraauthor{Sjoerd Dirksen}
\extraaffil{Mathematical Institute, Utrecht University, the Netherlands}
\extraauthor{Maurice Schmeits\correspondingauthor{Maurice Schmeits, Maurice.Schmeits@knmi.nl}}
\extraaffil{Royal Netherlands Meteorological Institute (KNMI), De Bilt, the Netherlands}

\abstract{\sjoerd{Current statistical post-processing methods for probabilistic weather forecasting
are not capable of using full spatial patterns from the numerical weather prediction (NWP) model. In this paper we incorporate spatial \maurice{wind speed} information by using convolutional neural networks (CNNs) and obtain probabilistic wind speed forecasts \maurice{in the Netherlands} for 48 hours ahead, based on KNMI's deterministic Harmonie-Arome NWP model.} \maurice{The probabilistic forecasts from the CNNs} \simon{are shown to have higher Brier skill scores for medium to higher \kiri{wind speeds}, as well as a better \maurice{continuous ranked \kiri{probability} score (CRPS) and logarithmic score}, than the forecasts from fully connected neural networks and quantile regression forests. As a secondary result\sjoerdr{,} we have compared the CNNs using 3 different density estimation methods (quantized softmax (QS), kernel mixture networks\sjoerdr{, and} fitting a truncated normal distribution), and \sjoerdr{found} the probabilistic forecasts based on the QS method to be best.}}

\begin{document}

\maketitle

\section{Introduction}
Accurate and reliable weather forecasts are important in many branches of society. Decision making in, for example, agriculture, aviation\sjoerd{,} and renewable energy production are all dependent on \sjoerd{skillful} weather forecasts \kiri{(e.g. \citet{wilczak2015wind})}. Furthermore, extreme weather can be dangerous \kiri{for life and property,} and it is therefore important to give reliable warnings when dangerous weather can be expected. \kiri{Extreme winds have a large impact in the Netherlands. Two western European wind storms caused one insurance group to pay an estimated €100 million \mauricer{due to} damages to individuals, businesses\sjoerdr{,} and the agricultural sector (https://news.achmea.nl/achmea-pays-out-over-eur-100-million-to-customers-hit-by-january-storms/). KNMI issued eight code orange weather warnings in 2018, of which \mauricer{at least three} were associated with extreme wind speeds, including the two winter storms that caused the damage noted above (https://www.knmi.nl/kennis-en-datacentrum/uitleg/archief-code-oranje-rood-in-2018 (in Dutch))}.

%Forecasts are generally produced by numerical weather prediction (NWP) models, such as \sjoerd{the Harmonie-Arome model (\cite{bengtsson2017harmonie}) of the Royal Netherlands Meteorological Institute (KNMI)}. Due to a lack of measurements a perfect initialization of these models is not possible. Simplifying assumptions have to be made to make computation feasible, but this parametrization of the sub-grid scale processes can introduce a bias in the forecast. Together these effects lead to errors in both the initialization and the forecast of the model. Besides, a single model run only provides a single deterministic forecast. The atmosphere is however a famously chaotic system (\cite{lorenz1963deterministic}) and every forecast is therefore inherently uncertain. A single forecast given by a NWP model does not provide an estimate of this uncertainty, even though it is important for decision makers to have such an estimate.
Forecasts are generally produced by numerical weather prediction (NWP) models, such as \sjoerd{the Harmonie-Arome model \citep{bengtsson2017harmonie} of the Royal Netherlands Meteorological Institute (KNMI)}. \sjoerd{To make computation of NWP models feasible it is necessary to make simplifying assumptions}, but \sjoerd{the resulting} parametrization of the sub-grid scale processes  can introduce \maurice{errors} in the forecast. \sjoerd{In addition, a perfect initialization of these models is not possible. %due to a lack of measurements. 
As the atmosphere is a famously chaotic system \citep{lorenz1963deterministic}, every forecast is therefore inherently uncertain. A single forecast given by an NWP model does not provide an estimate of this uncertainty, even though such an estimate is important for decision makers.} 
%Together these effects lead to errors in both the initialization and the forecast of the model. 
%Besides, a single model run only provides a single deterministic forecast. 

Forecast uncertainty is usually estimated from an ensemble of predictions where each member is \sjoerd{the outcome of an NWP model run} with a perturbed initial state and/or perturbed physical parameterizations. This \sjoerd{approach is}, however, computationally expensive and the results are often still biased and underdispersed \citep{Gneiting248}.

\sjoerd{To correct biases \mauricer{and systematic} errors in the ensemble spread\kiri{, and to make probabilistic forecasts from deterministic NWP forecasts,} one can use statistical post-processing}, based on past observations. A popular framework for \sjoerd{statistical post-processing} is model output statistics (MOS; \cite{glahn1972use}). In MOS a statistical relationship is derived between the \sjoerd{forecasts provided by the NWP model} and the corresponding \sjoerd{observed} measurements. In this way we can correct the bias and estimate the uncertainty in the forecast, based on  \kiri{deterministic or ensemble} \sjoerd{NWP} model output and potentially some \sjoerd{additional} variables, \sjoerd{such as} the time of the year. 
%We need a dataset containing both measurements and forecasts to apply MOS to model outpout. 

\kiri{It is common practice in statistical post-processing to fit a parametric distribution based on measurements and a set of potential predictors. These predictors can come from an ensemble forecast, as is the case in ensemble model output statistics (EMOS, \cite{Gneiting248}), or the predictors can come from a deterministic forecast in \sjoerdr{an} MOS application (e.g \cite{doi:10.1175/MWR-D-17-0290.1, bakker2019comparison}). \sjoerdr{In our study} we have used predictors from deterministic Harmonie-Arome model output as a long Harmonie-Arome \mauricer{ensemble} data set was not \mauricer{yet} available.} \sjoerd{The quality of the fit is measured in terms of skill scores associated with scoring rules such as the continuous ranked probability score (CRPS, \cite{MaW76,Her00,gneiting2007strictly}).} \kiri{Wind speed has been modelled with a parametric distribution in an }EMOS \kiri{framework} in e.g. \cite{scheuerer2015probabilistic}, \cite{thorarinsdottir2010probabilistic}, \sjoerd{and} \cite{doi:10.1002/env.2380}, where they used truncated normal and log\sjoerd{-}normal distributions.
Furthermore in \cite{doi:10.3402/tellusa.v65i0.21206} a mixture of truncated normal and the generalized extreme value distribution was used with success. \cite{ioannidis} tested a variety of distributions for wind speed in Denmark and found that the truncated normal distribution was the most \sjoerd{skillful}. 

%\kiri{These methods can be applied whether the covariates are from an ensemble prediction system or a deterministic NWP model, like we use here}. 

\kiri{Parametric methods (i.e.\sjoerdr{,} EMOS) have} been compared to quantile regression forests (QRF; \cite{meinshausen2006quantile}, a non-parametric technique based on random forests) for both wind speed and temperature forecasts in \cite{taillardat2016calibrated}, where QRF was found to be more \sjoerd{skillful}. QRF was also used in \cite{doi:10.1175/MWR-D-17-0290.1} for post-processing of precipitation forecasts\kiri{, and \citet{rasp2018neural} for post-processing 2m temperature forecasts}. 
\sjoerd{\cite{rasp2018neural} \mauricer{also } used fully connected neural networks (NNs) to determine the \mauricer{parameters of a normal distribution}. This approach was shown} to be more skillful than \kiri{fitting \mauricer{a normal} distribution in an EMOS framework} for the statistical post-processing of temperature forecasts. \sjoerd{In contrast to MOS/EMOS,} QRF and neural networks are both capable \kiri{of learning} non\sjoerdr{-}linear \kiri{relationships, while for MOS/EMOS non-linear dependencies must be specifically coded.} 

\sjoerd{The aforementioned methods (\kiri{MOS/EMOS}, QRF\sjoerdr{,} and fully connected neural networks) are not well suited to use high-dimensional structured spatial data, \kiri{although they can use spatial information in a crude way, for example by taking statistics of the predictor \mauricer{in an area} around the station location \citep{vanderplas2017comparative}.} As weather forecasts are spatial in nature, it could be beneficial to use post-processing methods that are capable of dealing with this spatial information. In the recent literature on machine learning, convolutional neural networks (CNNs) have strongly advanced the state-of-the-art on learning tasks involving this type of information, e.g., in image classification and time series analysis (see, e.g., \cite{lecun2015deep}, \cite{krizhevsky2012imagenet}). CNNs can potentially be of great benefit in the geosciences \citep{article} and have already been applied in \kiri{in the meteorological domain}. For example, \citet{DBLP:journals/corr/LiuRPCKLKWC16} used CNNs to detect extreme weather events in climate datasets, and \citet{shi2017deep} \sjoerdr{used} a mix between a convolutional and a recurrent network \sjoerdr{for} nowcasting precipitation. Additionally, \kiri{CNNs have been used to make statistical forecasts of frontal systems \citep{lagerquist2019deep}, 500 hPa geopotential height anomalies \citep{weyn2019can}, the probability of large hail using features from \sjoerdr{an} NWP model \citep{10.1175/MWR-D-18-0316.1}, and to estimate the uncertainty in weather forecasts based on the state of the atmosphere in the initialization of \mauricer{a} NWP model \citep{scher}.}

The above studies demonstrate the wide variety of problems for which CNNs have been used. However, to the best of our knowledge, CNNs have not yet been used for probabilistic forecasting \maurice{of wind speed} using statistical post-processing. We expect that the capability of CNNs to analyze spatial information of weather forecasts could make them a very beneficial new tool for this purpose. Independently of our work, \cite{10.1175/MWR-D-20-0096.1} very recently investigated CNNs for probabilistic forecasting of precipitation \kiri{on the subseasonal time scale} in California and found them to improve over state-of-the-art post-processing methods, and \sjoerdr{\cite{dupuy2020arpege}} demonstrated that CNNs are more skilful than traditional methods (QRF and logistic regression) in post-processing cloud cover forecasts.}

In this study we apply convolutional neural networks for the post-processing of +48h \kiri{deterministic wind speed} forecasts in the Netherlands. We compare \sjoerd{three different} methods for fitting a \sjoerdr{(conditional)} probability distribution using \sjoerdr{CNNs}\sjoerd{: quantized softmax \citep{oord2016pixel}, kernel mixture networks \citep{ambrogioni2017kernel}, and fitting a truncated normal distribution \cite[e.g.][]{thorarinsdottir2010probabilistic}} \sjoerdr{whose parameters are determined by the network}. Furthermore, we examine whether convolutional neural networks \sjoerd{are more skillful} than fully connected neural networks and QRF.

This paper is structured as follows. \sjoerd{In section 2 we give a description of the data that \maurice{has been} used in this study and in section 3 we give a short description of quantile regressions forests and \maurice{(convolutional)} neural networks and detail the models used in this study. Section 4 contains the results and, finally, section 5 contains the conclusions and discussion.} 

\section{Data}
The input data is provided by Harmonie-Arome cycle 40 (HA40) used by KNMI. HA40 is a non-hydrostatic model that is run on a 2.5 x 2.5 km grid. We \sjoerd{use} deterministic HA40 forecasts that are initialized at 0000UTC and are valid at a lead time of 48 hours. 
The predictand data are the 10-minute-average wind speed observations in the extended winter period (mid-October to mid-April), at 10 meters above the ground, from \maurice{46} weather stations in the Netherlands, which are shown in Table \ref{tab:loc_weather}. These measurements are provided as rounded to the nearest m/s. The data from all the stations are pooled in the training dataset, meaning that the model is trained for all stations at once, without providing station\sjoerdr{-}specific information other than \mauricer{HA40}  surface roughness. 

Reforecast data for HA40 is available from 2015 until 2017 and operational \mauricer{HA40} forecasts from winter 2018-2019 are \mauricer{also} available. This data \sjoerd{is} split into two sets, as shown in Table \ref{tab:folds_}. The first set (2015-2017) is used for model selection and training. We use a three-fold cross-validation on this model selection set. The second set (2018-2019) is an independent data set used for testing the selected models.

In three-fold cross-validation we \sjoerd{train} every model three times on the model selection set, \sjoerd{each time} with a different fold left out. The latter \sjoerd{is} then used to make predictions in order to validate the model. The sets are chosen in this way to ensure that there is at least six months between the training, \sjoerdr{validation, and} test sets. This is necessary to avoid temporal correlations between the different data sets. 

%\subsection{Predictors}
In this study we use two sets of predictors. The first set contains the HA40 forecasts of a number of variables in the neighbourhood of the station\sjoerdr{, see Table~\ref{bo: pred1}}. The second set contains the wind speed forecast from HA40 for a larger area around this station, \simon{the exact size of which has been determined in the hyperparameter search}. The first set is used in all the methods described. The second set is only used for convolutional neural networks \sjoerd{(in combination with the first set)}.

The set with the neighbourhood predictors we use in this study is based on previous research on post-processing of wind speed forecasts by \cite{ioannidis} and \cite{taillardat2016calibrated}. Based on their results we take the variables shown in Table \ref{bo: pred1} as the set \sjoerd{of potential predictors}. %\sjoerd{Variables are selected from this set via a greedy algorithm which adds predictors successively based on which predictors reduce the mean squared error (MSE) the most.\simon{This part talks about linear regression so i commented it out}}

%The grid point closest to the station is used for the surface roughness. For the other variables we \sjoerd{pick} a number of gridboxes around the station and \sjoerd{determine} the mean value, maximal value and minimal value of each predictor in this region, \sjoerdr{so that the method receives} some information about the spatial uncertainty in the \sjoerdr{weather} forecast. For the  surface roughness only the closest grid-point has been taken since the spatial uncertainty does not exist for this parameter.
\sjoerdr{The grid point closest to the station is used for the surface roughness. For the other variables we pick a number of grid boxes around the station and determine the mean value, maximal value, and minimal value of each predictor in this region, so that the method receives some information about the spatial uncertainty in the weather forecast. For the surface roughness only the closest grid-point has been taken since it should reflect the conditions in the direct neighbourhood of the station as closely as possible.}
Which predictors we will use, the number of gridboxes used, and whether to take the mean, maximum, minimum or a combination of them is decided for every method independently in the hyperparameter search, which is described in section~\sjoerdr{\ref{sec:methods}}\ref{ref: section hyperparameter}. 

\section{Methods}
\label{sec:methods}

Three different methods \sjoerd{are} compared in this study: quantile regression forests, fully connected neural networks and convolutional neural networks. We \sjoerd{also compare} three different methods \sjoerd{for conditional density estimation} using convolutional neural networks. \sjoerd{Some of the models are trained by using the errors of linear regression as target variables instead of the observed measurements. We will motivate this choice for the various models below.}

\subsection{Quantile Regression Forests}
Quantile regression forests \sjoerd{\citep{meinshausen2006quantile} is} a non-parametric method for \sjoerd{estimating quantiles and, more generally, a conditional cumulative distribution function}.
The algorithm is based on random forests (\cite{breiman2001random}). \sjoerd{Whereas a trained random forest outputs a point prediction by taking the average of the terminal nodes, QRF returns an estimate of the cumulative distribution function}. This algorithm was shown to outperform EMOS methods for post-processing of both \kiri{wind speed} and temperature forecasts by \cite{taillardat2016calibrated} and for precipitation by \cite{doi:10.1175/MWR-D-17-0290.1}, and will therefore be used as a benchmark in this research. 

We \sjoerd{use} the Python package \sjoerd{Scikit-garden} to implement quantile regression forests. 
%(\cite{scikit-learn}). 
Within this package there is no option to obtain a full \sjoerd{cumulative distribution} function. Therefore an alternative prediction function \sjoerd{is} used which outputs the average of the empirical cumulative distribution functions of the leaves of every tree in the random forest.

For quantile regression forests the most important \mauricer{hyperparameters}  \sjoerd{are} the minimum leaf size of the trees and the amount of randomization. We can control the \sjoerd{amount of} randomization in the random forest by \sjoerd{varying the size of the} random subset of predictors \sjoerd{that is used} for splitting at every step. Other hyperparameters that are explored are the \sjoerdr{the choice of the} impurity function and the number of trees. 
%For the latter we used 100 trees in the first hyperparameter search and 500 for the final model. For the impurity function we compared the mean squared error to the mean absolute error, and found the former to give the best results.

We \sjoerd{train} quantile regression forests \mauricer{using either the observations or} the residuals of linear regression. The second approach could be beneficial for two reasons. \sjoerd{Firstly,} quantile regression forests cannot extrapolate outside the range of the training data. \sjoerd{As linear regression is able to extrapolate, we may be able to obtain a better model for higher wind speeds by combining QRF and linear regression.} \sjoerd{Secondly,} random forests split the data into boxes based on which split minimizes the total impurity. \sjoerd{If the relationship between the response variable and a single predictor is linear, then} it may take random forests many splits to represent this relationship. Splits based on other variables are as a \sjoerd{result} made with limited information. Fitting to the residuals of a linear model can reduce this effect.

\subsection{Conditional \sjoerdr{D}ensity \sjoerdr{E}stimation using Neural Networks}
\label{sec:densEstNNs}

In this work we consider three different methods for conditional density estimation using \mauricer{convolutional} neural networks. These methods are quantized softmax, kernel mixture networks with Gaussian kernels, and parametric density estimation with a truncated normal distribution. The first method adds an additional quantized softmax output layer to the neural network to create an estimate of the conditional probability density function by a histogram with pre-defined bins. \mauricer{This method is also used for conditional density estimation using fully connected neural networks}. The second method fits a mixture of normal distributions, where the mean of every Gaussian is fixed but the weights \sjoerdr{in the mixture} and \sjoerdr{the} standard deviations \sjoerdr{of the Gaussians} are learned by the network. \simon{The means of the Gaussians are taken to lie on a regularly spaced grid between -15 and 15 m/s, i.e. the set of means \mauricer{of the residuals} is given by $\{-15,-15+\frac{30}{N},...,-15+30\frac{N-1}{N}\}$. The number of kernels $N$ is used as a hyperparameter in this study.} The third method fits a truncated normal distribution. In this case the network learns the two parameters of the distribution. In case the network is trained directly on observations, the normal distribution is truncated at zero. If the network instead trains on residuals, then the distribution is truncated at minus the forecast of the linear regression in order to ensure that negative \kiri{wind speeds} cannot be predicted. In the Appendix a more detailed description of the three methods is given.

We train the neural networks by empirical loss minimization. As potential loss functions we consider the continuous ranked probability score (CRPS) and the negative log\sjoerdr{-}likelihood. The CRPS of a given conditional cumulative distribution function estimate $\hat{F}$ (associated with a conditional probability density function estimate) and a training datum $(x,y)$ is defined by
\begin{equation*}
\operatorname{CRPS}(\hat{F},(x,y)) = \int_{-\infty}^{\infty} (\hat{F}(c|x)-\ind_{[y,\infty)}(c))^2 dc.
\end{equation*}
The \mauricer{log(arithmic) score is defined as the} negative log-likelihood of a given conditional density function estimate $\hat{p}$ and training datum $(x,y)$\mauricer{, and} is given by 
\begin{equation*}
\mathcal{L}(\hat{p},(x,y)) = -\log(\hat{p}(y|x)).
\end{equation*}

\subsection{Fully Connected Neural Networks}

In this section we give a concise description of the fully connected neural networks used in this work. For a general introduction to neural networks and the standard terminology used in this work we refer to \cite{Goodfellow-et-al-2016}. We explore networks whose first part is a stack of $n$ blocks\sjoerdr{, each of which contains a dense layer} of size $m$ followed by a ReLU activation function and a dropout layer. We \mauricer{have} also \sjoerdr{explored} architectures with batch normalization layers, \sjoerdr{as are} used in the convolutional neural networks \sjoerdr{(see section~\ref{sec:methods}\ref{sec:MethodsCNNs})}, but \sjoerdr{found that these did not} increase the performance.

For the fully connected neural networks we use the quantized softmax method for conditional density estimation. We minimize the empirical loss associated with either the CRPS or the negative log-likelihood using adaptive moment estimation (Adam; \cite{kingma2014adam}), a variant of stochastic gradient descent that is very popular in deep learning. We use early stopping to determine the number of epochs (the number of times the training data is \sjoerdr{used during training}).

The neural networks used in this research were programmed using Keras (\cite{chollet2015keras}), with TensorFlow as backend (\cite{tensorflow2015-whitepaper}). Adam was used using default options for all parameters other than the learning rate \sjoerdr{decay parameter}.

As in the case of QRF, the fully connected neural network is trained \mauricer{using either the observations or} the residuals of linear regression. For neural networks applying linear regression is hypothesized to give better results due to the fact that lower wind speeds are much more prevalent in the training data set. Output neurons which are related to high wind speeds therefore need to be activated in only a very small sample of the data. In case of direct training, we use a softmax layer with $30$ output bins, where every bin (of size 1 m/s) represents a different wind speed ranging from 0 to \mauricer{30} m/s. For the neural network which is trained \mauricer{on} the residuals of linear regression as target variables, we use $300$ \sjoerdr{identically sized} output bins. In this case, every \sjoerdr{bin} represents a different value of the residual ranging between $-15$ and $15$ m/s. In the linear regression case we use a higher resolution as the errors of linear regression can take on any value, whereas the actual measurements are rounded to the nearest m/s. When training on the residuals, we add Gaussian noise with mean zero and variance $\sigma^2$ to the target variables to smoothen the results. \sjoerdr{This} is necessary \sjoerdr{as} the number of bins is rather large compared to the number of values in the training dataset.

The hyperparameter search is performed on the number of \sjoerd{blocks} $n$ and layer size $m$, dropout rate, \sjoerd{$\ell_1$-}regularization strength, learning rate, batch size and, in case we use linear regression, the label noise variance $\sigma^2$. Moreover, we investigate the effect of using the CRPS and the \sjoerdr{negative} log-likelihood during training. We furthermore check the same potential predictor variables as for QRF \mauricer{(Table~\ref{bo: pred1})}. \sjoerdr{Finally, we used the learning rate decay parameter of Adam as a hyperparameter}.
%\simon{We also used the learning rate decay parameter as a hyperparameter, this is not necessary when you use ADAM as an optimizer since this method changes the learning rate by itself. The hyperparameter was however used and is therefore been mentioned for completeness sake.}

\subsection{Convolutional Neural Networks}
\label{sec:MethodsCNNs}

When applying neural networks to high\sjoerd{-}dimensional input data, such as images, the number of trainable parameters becomes very large. 
Convolutional neural networks are neural networks which are specialized in analyzing images, by limiting the number of parameters in the network based on the structure of the task at hand. This is \sjoerdr{achieved} using two techniques. The first technique is parameter sharing\sjoerdr{: a} number of parameters are assigned the same value \sjoerdr{to ensure} that the same transformation is applied everywhere in the input image. \sjoerdr{In object detection tasks, this ensures that it is irrelevant where the object we want to detect is located in the image}. The \sjoerdr{second} technique is local \sjoerdr{connectivity, i.e.,} only connecting neurons which are related to pixels from an input image which are close to each other. This is based on the assumption that the relation between pixels that are close to each other is important.
%This assumption is ofcourse not a completely valid one in meteorology, but the appearance of certain structures in the weather forecast might still help in post processing even when the exact location is unknown to the model.  

%The two techniques used to limit the number of parameters in a convolutional layer are parameter sharing, which ensures a degree of translation invariance, and local connectivity, which corresponds to the gridded nature of images.

\sjoerd{For the same reason as for fully connected neural networks, we investigate training of CNNs on the residuals of linear regression. An additional motivation,} which is more specific \sjoerd{to} CNNs, is that we can use local information for linear regression. \sjoerd{The strength of a} CNN is based on the translation invariance of the \sjoerdr{spatial} patterns \sjoerdr{that} it needs to learn. \sjoerd{The wind speed at a particular weather station is, however, expected to be strongly dependent on the wind speed forecast of the NWP model at that station.} The translation invariance of the convolutional layers is therefore not suited for predictions at a specific weather station. Features in the forecast that correlate to the bias and the \sjoerd{forecast uncertainty are expected to be less local in nature and should therefore be better suited for analysis using CNNs}.

The CNNs \sjoerd{all receive} two different inputs. The first input is the full spatial forecast of the wind speed for a certain region around the weather station, which provides the corresponding observation. This is the input which is \sjoerd{received by} the convolutional part of the network. The second input contains the other variables, \simon{obtained} from the nearest grid boxes around the station, similar to what is \mauricer{used} for QRF and the fully connected neural networks.

\mauricer{The final architecture of the network is shown in Table~\ref{tab: conv_architecture}}. The convolutional part of the network consists of $n_{conv}$ layers with $m_{conv}$ filters, \simon{where the same number of filters is used in every layer}. Each of these \sjoerd{convolutional} layers is \simon{followed by a Relu activation function, a batch normalization layer and a max pooling layer}, where we use a step size of 2 by 2 for the Max Pooling layer and a filter size of 3 by 3 for the convolutional layer. For the fully connected part of the network \simon{every dense layer was followed by a Relu activation function, a batch normalization layer and a dropout layer}. 
 
For the convolutional neural network we compare all three methods for \sjoerd{conditional density estimation} that were discussed in Section~\ref{sec:methods}\ref{sec:densEstNNs}, i.e., quantized softmax, kernel mixture networks, and \sjoerdr{fitting} a truncated normal distribution. We again train the networks by minimizing the empirical loss associated with either the CRPS or the negative log-likelihoood using Adam. 

The size of the output layer of the convolutional neural networks depends on which conditional density estimation method is used. For quantized softmax, from here on referred to as CNN\_LR, the output layer has size $300$, as is also used for NN\_LR.
For \sjoerdr{fitting a} truncated normal, from here on referred to as CNN\_LR\_N0, we \sjoerdr{use} two output neurons \sjoerdr{(corresponding to the two parameters of the distribution)} and for the kernel mixture network, from here on refered to as CNN\_LR\_KMN, we need two output neurons for every kernel that we use. 

\subsection{Predictor and hyperparameter selection}
\label{ref: section hyperparameter}

Due to the large number of different parameters and potential predictor variables, \sjoerdr{it is computationally infeasible to select variables and hyperparameters using} a full grid search. To select predictors, model architectures, and hyperparameters we use a random search \sjoerdr{through a large range of options} for both the neural networks and the random forests. \sjoerdr{Based} on the results of a large sample of models we \sjoerdr{narrow} down the range for hyperparameters that are most important and repeat the procedure. In this \sjoerdr{way} we can search through a large space of possible models with \sjoerdr{limited} computational resources. \sjoerdr{The initial range of the hyperparameters is selected large enough to ensure that models with parameter values on the high and low end of the range perform clearly worse than models with intermediate parameter values. In this, we enlarge the chances that good parameter values are contained in the initial range. The only exception to this rule is the selection of the batch size for the CNNs: in this case we checked batch sizes that are powers of 2 and select the largest batch size that is computationally feasible for the computer used.}
%This methodology naturally tunes the more important parameters such as the necessary predictors, the model size and the learning rate first.  

\section{Results}
\label{sec:Results}

\subsection{Variable selection and hyperparameter tuning}

\sjoerd{As has been explained above, some models have been trained on the errors of linear regression instead of on the measurements directly. In these cases, linear regression was fitted on the mean values \kiri{of 10m wind speed, surface roughness and 925 hPa meridional and zonal wind components (predictor variables 2, 3 and 4 as defined in Table \ref{bo: pred1})} on an area of 12.5 by 12.5 km around the stations. These variables were selected through \simon{forward stepwise selection, }a greedy algorithm which adds predictors successively based on which predictors reduce the mean squared error (MSE) the most. As the MSE did not improve significantly after these predictors had been selected, all other candidate variables have been left out. However, all candidate predictor variables have been used in the non-linear methods, as they improved results in all cases.

The best QRF models, as determined by the hyperparameter search, have the following characteristics. \sjoerdr{The predictor data contain the maximum, minimum and mean value of the predictors \kiri{ that are} \sjoerdr{marked} in bold \sjoerdr{in} Table \ref{bo: pred1}.} The best results were obtained by using this full set of predictors \sjoerdr{for splitting at every step}, so that decorrelation between the trees only occurs through bootstrapping on the training set. For the impurity function we compared the mean squared error to the mean absolute error, and found the former to give the best results. For the random forest trained on the wind speed measurements, hereafter referred to as QRF, we have used a minimum leaf size of 30. For the random forest trained on the residuals of linear regression (QRF\_LR), we have used a minimum leaf size of 42.
Oversampling the data, such that training samples corresponding to high wind speed days are shown to the \simon{random forest} more often during the training phase, was tried, but this appeared to have a negative impact on the results. This may be due to the fact that this leads to a large number of copies of outliers in the training set which do not generalize well. \simon{Furthermore oversampling based on \mauricer{observations gives a bias for higher windspeed values, and therefore oversampling based on HA40 wind speed forecasts would} probably have been a better choice.} Less naive oversampling methods with data augmentations might be more useful still, but were not tried. We used 100 trees during the first hyperparameter search and 500 trees for the final model.

The neural network trained on the wind speed measurements themselves, hereafter referred to as NN, appears to give the best results if it uses the maximum and mean value \kiri{of the sine and cosine of 10m wind direction, 10m wind speed, and surface roughness (predictor variables 1, 2 and 3 from Table \ref{bo: pred1})}. 
The neural network trained on the residuals, hereafter referred to as NN\_LR, gives the best results when trained on the means of the \kiri{bold} predictor variables shown in Table \ref{bo: pred1} and the maximum and minimum value of the wind speed. $\ell_1$-regularization did not appear to improve the results and was left out completely for both methods. The values of the other hyperparameters are shown in Table \ref{tab:nn}.

The convolutional networks have all been trained on the residuals of linear regression. 
%We refer to the three final networks with the quantized softmax, truncated normal, and kernel mixture network as CNN\_LR, CNN\_LR\_N0, and CNN\_LR\_KMN, respectively. 
Convolutional neural networks trained on the observations were found to be not \sjoerd{skillful} in preliminary testing. This was partly due to the fact that networks trained on the observations directly took longer to converge and converged to poor values more often than models trained on the residuals. This resulted in a significantly slower hyperparameter search. No real difference in performance is observed between the CRPS and log-likelihood as loss functions, neither in training time nor in the final result.}
\simon{The log-likelihood is, however, more sensitive to the initialization, since a poor initial estimate leads to exploding gradients. This is less of an issue when using the CRPS, since for a deterministic forecast the CRPS is equal to the mean absolute error, for which the derivative is piecewise constant. This results in a more stable behavior during the training phase. 

The hyperparameters used in the hyperparameter search and the selected values of each of these hyperparameters are shown in Table \ref{tab:conv_hyper}.}

\subsection{Verification results for models trained on \kiri{two-thirds} of the training data}
\label{sec:ResultsVali}

The CRPS results for the three different cross-validation folds for the best models of all methods are shown in Table \ref{table;results}. These results show that convolutional neural networks outperform QRF and NN on all three folds in cross-validation. Hyperparameters were selected based on these results however, \sjoerd{and therefore we have also checked the CRPS on the independent test set (Table~\ref{tab:results}). On the latter set,} three different forecasts were verified using every method. Each of these forecasts is based on the model trained on a different training set as used in the cross-validation\sjoerd{, in order} to obtain an estimate of the variation in the results when different training data is used. 
\sjoerd{A downside of this procedure is that it may favor} neural networks over quantile regression forests, due to the fact that we use early stopping based on the validation set \sjoerdr{in training the neural networks}. \sjoerd{Therefore a final comparison is made between the CNNs and QRF when they are trained on the full training dataset} (subsection~4\ref{subsect:trainfull}).

\begin{figure}[t!]
    \centering
    \begin{subfigure}[t]{0.45\textwidth}
        \centering
        \includegraphics[scale = 0.45]{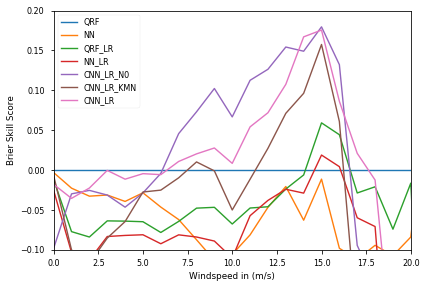}
        \caption{\mauricer{BSS for fold 3}}
    \end{subfigure}%
    \begin{subfigure}[t]{0.45\textwidth}
        \centering
        \includegraphics[scale = 0.45]{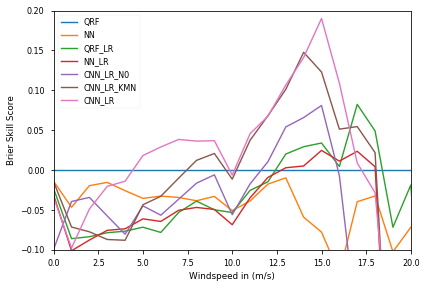}
        \caption{\mauricer{BSS for fold 1}}
    \end{subfigure}
        \begin{subfigure}[t]{0.45\textwidth}
        \centering
        \includegraphics[scale = 0.45]{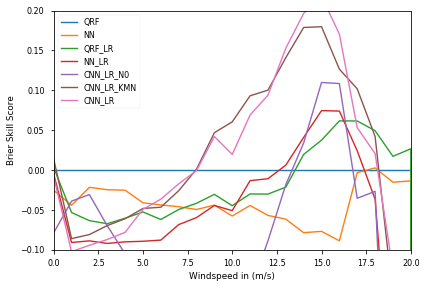}
        \caption{\mauricer{BSS for fold 2}}
    \end{subfigure}
    \begin{subfigure}[t]{0.45\textwidth}
        \centering
        \includegraphics[scale = 0.45]{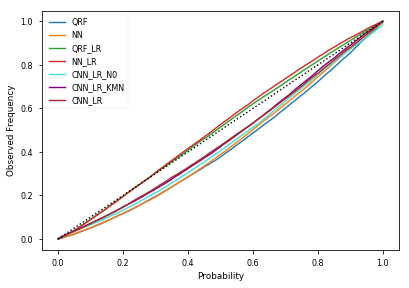}
        \caption{\mauricer{PIT diagram}}    
    \end{subfigure}%
\caption{(a-c) Brier skill scores of the different methods relative to QRF, for predictions trained on three different training sets \maurice{(see \mauricer{Table \ref{tab:folds_})}}. (d) \mauricer{PIT diagram} \maurice{for the forecasts of the different methods for the 3 cross-validation sets combined}.}
\label{fig: Brier_three_folds}
\end{figure}

In Table \ref{tab:results} the results are shown for the root mean squared error, the mean absolute error, the CRPS, and the log score based on the independent test set. These results show that adding spatial information through convolutions reduces the error of both the deterministic forecast (i.e., \sjoerd{the} mean and the median of the probabilistic forecast for the RMSE and MAE, respectively) and the probabilistic forecast. Furthermore we can see that applying linear regression improves the %deterministic forecast of both QRF and 
\mauricer{scores of the} fully connected neural networks. \sjoerd{However}, it does not improve the \mauricer{scores of QRF, except for the RMSE}. 
In Figure \ref{fig: Brier_three_folds}(a-c) the Brier skill score relative to QRF is shown for the three different training sets. From this \maurice{figure} it is clear that convolutional neural networks are more skillful than the other methods at higher wind speeds. For wind speeds above 18 m/s their performance becomes worse again, however in this range there is not enough data \sjoerd{to draw} any conclusions.
Figure \ref{fig: Brier_three_folds} also shows that learning the residuals of linear regression mainly helps to improve forecasts for higher wind speeds, while for low wind speeds the results become worse for both neural networks and random forests.

Figure \ref{fig: Brier_three_folds}(d) shows the probability integral transform (PIT) \mauricer{diagram} of all the methods. In this figure we can see a clear difference between the models trained on the wind speed observations and models trained on the residuals of linear regression for QRF and the fully connected neural networks. \sjoerd{The} methods trained on the residuals lie closer to the diagonal, implying that on average they make \sjoerd{a better estimate of the }\simon{probabilities}. It is surprising, however, that this does not hold for the CNNs which are trained on residuals. For QRF and the CNNs we see that the \sjoerdr{PIT curve} lies under the diagonal, which \sjoerd{means} that for these methods observations fall in the higher quantiles of the estimated distribution more often than expected. This implies that the probability of higher \kiri{wind speeds} is underestimated by these methods. 

\subsection{\maurice{Verification} results for models trained on the full training dataset}\label{subsect:trainfull}

\begin{figure}[t]
    \centering
    \begin{subfigure}[t]{0.5\textwidth}
        \centering
        \includegraphics[scale = 0.5]{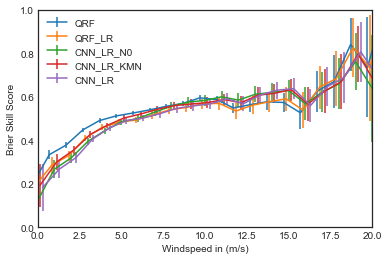}
    \end{subfigure}%
    \begin{subfigure}[t]{0.5\textwidth}
        \centering
        \includegraphics[scale = 0.5]{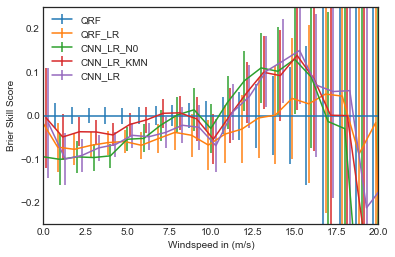}
    \end{subfigure}
    \caption{Brier skill scores relative to the station climatology (left) and QRF (right), for models trained on the full training data set. \sjoerd{The error bars represent the estimates of the standard deviation} obtained by \maurice{block} bootstrapping the test data 1000 times. Block bootstrapping was used to \sjoerd{ensure that} spatial correlation between stations is accounted for.}
\label{fig: bss full}
\end{figure}

\sjoerd{We make a final comparison of QRF and the CNNs by training the models on the entire} training dataset; fully connected neural networks are omitted since they \sjoerd{showed poorer performance in} the results of the previous subsection. For the convolutional neural networks the number of epochs was chosen to be 2/3 of the average number of epochs that gave the best results in cross-validation, i.e., 6, 12, and 16 for CNN\_LR\_N0, CNN\_LR\_KMN and CNN\_LR, respectively. This \sjoerdr{choice ensures that} the number of training steps \sjoerdr{is the same as in Section~\ref{sec:Results}\ref{sec:ResultsVali}}.
The results obtained for the CNNs \simon{are comparable, in terms of CRPS \mauricer{(Table~\ref{table: results_fullset})}, to the case where they were trained on only part of the training data \mauricer{(Table~\ref{tab:results}). We see in Table~\ref{table: results_fullset} that the CNNs still outperform QRF}. The BSS of the CNNs, compared to QRF,} \sjoerdr{is slightly lower, however, for wind speeds between 12} \mauricer{and 16 m/s (cf. Figs.~\ref{fig: bss full} and \ref{fig: Brier_three_folds}(a-c))}.

%\sjoerd{are slightly worse than in the case where they were trained} on only a part of the training data. This \sjoerd{could be} due to the fact that for the latter networks we could use the validation dataset \sjoerd{for early stopping}, \sjoerd{thereby increasing the generalization performance of the networks}. 

%\sjoerd{The results of the comparison} are shown in Table \ref{table: results_fullset}. We see that the CNNs still outperform QRF.
Figure \ref{fig: bss full} shows the Brier skill scores of the models trained on the full data set with respect to both the station climatology (left panel) and QRF (right panel). \sjoerd{In this figure we include a bootstrap estimate of the standard deviation}, obtained by block bootstrapping 1000 times\maurice{, i.e., by drawing data from all stations of a single date at once because of spatial correlation}. From this figure it is clear that the CNNs \sjoerd{perform} better than QRF for higher \kiri{wind speeds} \simon{($\sim{11}-15$ m/s)}. Furthermore, it is clear that for wind speeds above 15 m/s the uncertainty in the Brier skill scores is much larger than the difference in Brier skill scores between the methods. 

\begin{figure}
    \centering
    \includegraphics[scale = 0.55]{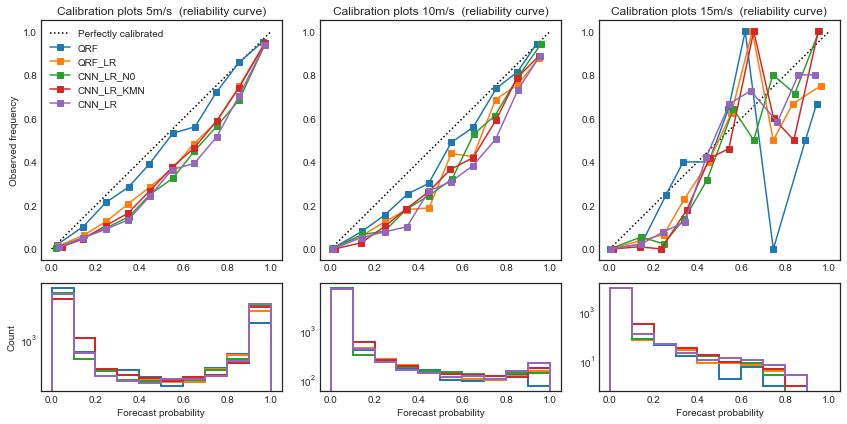}
    \caption{Reliability diagram for the CNNs and QRF, trained on the full training data set, \maurice{for thresholds of 5 m/s (left panel), 10 m/s (middle panel) and 15 m/s (right panel)}. \simon{The exceedance frequencies for these thresholds in our \mauricer{test} data set are  49.2\%, 9.3\% and 1.0\%, respectively.} }
    \label{fig:15}
\end{figure}

Figure \ref{fig:15} shows reliability diagrams for \mauricer{thresholds of} 5, 10 and 15 m/s. Here we can see that for 5 m/s the forecasts of QRF are better calibrated, but both the CNNs and QRF\_LR forecasts are somewhat sharper. For 10 m/s the QRF forecasts are still better calibrated, but they give \mauricer{less often} a high probability of exceeding this threshold and are slightly less sharp. Finally, for 15 m/s we can see that QRF is significantly worse at predicting these events when they are likely; both the calibration and sharpness are worse than for CNNs.

\subsection{Geographic differences \maurice{in CRPSS}}
\begin{figure}
        \centering
        \includegraphics[scale = 1]{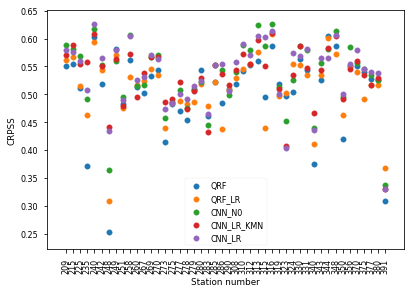}
        \caption{CRPSS with respect to station climatology of different methods based on the models trained on the full training data set. Station numbers are explained in Table \ref{tab:loc_weather}. }
        \label{fig: stations1}
    \end{figure}

Figure \ref{fig: stations1} shows the difference in continuous ranked probability skill score (CRPSS), with respect to the climatology, between QRF and the CNNs for the different stations. This shows that CNN is the most \sjoerd{skillful} method for almost all stations, although the CRPSS difference between the methods is relatively small for most stations. 

\begin{figure}[t]%{0.5\textwidth}
        \centering
        \includegraphics[scale = 1]{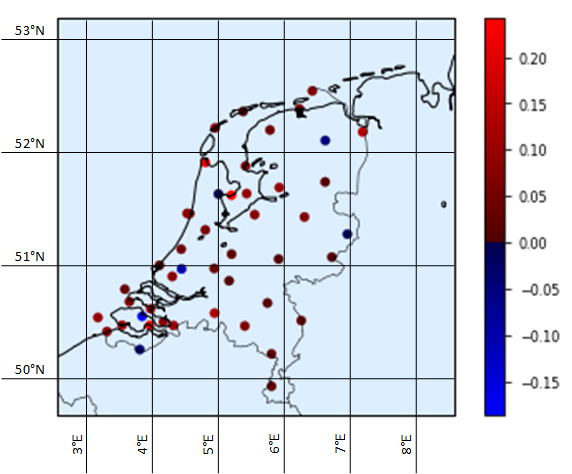}
\caption{CRPSS of CNN\_LR with respect to QRF. Here positive values imply that CNN\_LR is more \sjoerd{skillful} than QRF. }
\label{fig: stations2}
\end{figure}

\sjoerd{By visualizing the} activations of the convolutional layers for the CNNs one can observe that \sjoerdr{this} method is able to detect the Dutch coastline \sjoerd{(see \cite{veldkamp2020statistical} for details)}. Based on this \sjoerd{observation}, we could hypothesize that the CNNs are more \sjoerd{skillful} for higher wind speeds due to a higher skill for coastal stations. Figure \ref{fig: stations2} shows the CRPSS of CNN\_LR with respect to QRF on a map of the Netherlands. We cannot see \maurice{a clear indication of higher CRPSS values of CNN\_LR for coastal stations}, so that the higher skill of the CNNs is not fully explained by its ability to differentiate between coastal and non-coastal stations. 

\section{Conclusion and Discussion}
We have shown that for +48 hour wind speed forecasts convolutional neural networks can be of added value for statistical post-processing. Convolutional neural networks outperform \sjoerd{quantile regression forests} and fully connected neural networks, in terms of CRPS, in all the three cross-validation sets\mauricer{, and in terms of CRPS, log score, MAE and RMSE in} the final independent test set.
Besides, we have compared the CNNs using 3 different density estimation methods (quantized softmax (QS), kernel mixture networks\sjoerdr{, and} fitting a truncated normal distribution), and \sjoerdr{found} the probabilistic forecasts based on the QS method to be best.

The Brier skill score shows that CNNs outperform QRF for higher wind speeds that are more important in weather forecasting \kiri{because of their potential impact on society}. \sjoerd{In contrast}, for \kiri{wind speeds} up to $\sim{10}$ m/s QRF has both a better Brier skill score and is better calibrated. 
The \sjoerd{poor} performance of the CNNs with respect to QRF in the lower wind speed range could be explained as an effect of using ordinary least squares regression. \sjoerd{The latter assumes errors that are symmetrically distributed around zero and therefore} does not perform well \simon{for low wind speeds,} \sjoerdr{as this method} implicitly assumes that negative wind speeds are possible. This could be mitigated \sjoerd{by performing a variant of ordinary least squares that excludes this possibility}. For wind speeds above 15 m/s the uncertainty in the Brier skill score grows very fast and conclusions for this range can therefore not be \sjoerd{drawn}. This is mainly caused by a lack of \mauricer{cases} with high wind speeds in the available data set. \simon{The test set, for example, only includes a single measurement above 19 m/s.}
An obvious solution for this would be to obtain more data by obtaining reforecast data for more years, assuming these years contain more climatologically extreme wind speeds. A \sjoerd{less costly} solution to this problem could be to reforecast days in the past with more extreme weather, such as days on which weather warnings were issued, instead of reforecasting full years only, as is \sjoerd{currently} done.

\sjoerd{Although convolutional neural networks proved to be most skillful in our study, a drawback of this method is that it is difficult to interpret for a meteorologist. In the appendix of \cite{veldkamp2020statistical} one can find a few figures showing the activations in the convolutional layers of the network for a number of days. These do not give a clear indication of which input features are important, \kiri{although the coast line can be clearly seen}. In future research it would be a good addition to use explainability methods, such as layer-wise relevance propagation (\cite{bach2015pixel}), to visualize which parts of an input image are most relevant for the prediction made by a convolutional neural network. This could be especially useful when fitting a truncated normal distribution, as in this case it may be possible to distinguish between features that are relevant in correcting the bias and features that are relevant in predicting the spread.} In an ideal case, identifying features \kiri{that} \maurice{are able to correct} a large bias or \maurice{reduce the} spread might even help in identifying shortcomings in the NWP model.

\sjoerdr{Another drawback of convolutional neural networks is the fact that they require a large amount of training data and are therefore probably less suited for \mauricer{local (i.e. for each station separately)} post-processing. In this study we focused on training \mauricer{global models (i.e. for all stations at once)} without using station specific information apart from model surface roughness. This has the benefit that there is more training data available and that the resulting models can be used to post-process the weather forecast for all grid cells instead of only for specific stations. For countries with \mauricer{mountainous areas topographical predictors should be added in such a global model framework \citep[e.g.][]{rasp2018neural}}}.

At the time this study was conducted not enough data was available \sjoerd{from the} \maurice{KNMI} \kiri{Harmonie-Arome} ensemble forecasts. \sjoerd{As many} current \maurice{statistical} post-processing studies are based on ensemble output\sjoerd{, an} important next step would therefore be to investigate if convolutional neural networks also add skill when potential predictors are taken from ensemble forecasts.

\acknowledgments 
\simon{We would like to thank \maurice{the following people from KNMI: Toon Moene for executing the reforecasting runs of the Harmonie-Arome model, and} Andrea Pagani and Dirk Wolters for assisting with the practical implementation of the deep learning methods used in this paper. Besides, we are grateful to the 3 anonymous reviewers for their comments, which have helped to improve a previous version of the manuscript. S.D. acknowledges funding by the Deutsche Forschungsgemeinschaft (DFG, German Research Foundation) under SPP 1798 (COSIP - Compressed Sensing in Information Processing).}

\appendix
\appendixtitle{}

\sjoerd{In this appendix we briefly discuss the three methods that have been used to estimate the conditional probability density function \mauricer{of the neural networks}.}

\subsection{Quantized Softmax}
\sjoerd{Quantized softmax (\cite{oord2016pixel}) is a simple method to obtain an estimate for the conditional density using neural networks. The goal of the method is to approximate the conditional density by a histogram with $m$ predetermined bins $A_1,\ldots,A_m$. For this purpose we construct a neural network (with a linear output layer) with $m$ output neurons and apply the softmax function
\begin{equation*}
\operatorname{Softmax}(z)_{i}=\frac{e^{z_i}}{\sum_{j=1}^{m}e^{z_j}}, \qquad i=1,\ldots,m    
\end{equation*}
to the output of the last layer. We can turn the resulting output $w(x)$ (associated with an input datum $x$) into an estimate $\hat{p}(y|x)$ for the conditional probability density function by setting
\begin{equation*}
    \hat{p}(y|x) = \sum_{i=1}^{m}\frac{1}{\operatorname{Vol}(A_i)}\ind_{A_i}(y) w(x)_i
\end{equation*}
where $\operatorname{Vol}(A_i)$ is the \sjoerdr{size} of bin $A_i$. 
The set of probability densities that can be approximated well by this procedure is controlled by the choice of the bins $A_i$.}

\subsection{Kernel Mixture Networks}

\sjoerd{The second method used in this paper for conditional density estimation is the kernel mixture network (KMN; \cite{ambrogioni2017kernel}). We describe here directly the variant of KMN with Gaussian kernels, which is used in our study. This variant is very similar to a mixture density network (\cite{bishop1994mixture}). The kernel mixture network estimates the conditional density by a mixture of Gaussians in which the means are fixed and the weights and variances are learned by a neural network. Let $Y=\{y_1,\ldots,y_m\}$ be a subset of the label space containing the kernel centers. Let $\phi(y) = \frac{1}{\sqrt{2\pi}}e^{\frac{-y^2}{2}}$ denote the standard Gaussian density. We construct a neural network (with a linear output layer) with $2m$ output neurons. To the first $m$ outputs we apply the softmax function and denote the resulting output for \sjoerdr{a given} input datum $x$ by $w(x)$. The entries of $w(x)$ are the weights in the mixture of Gaussians. To each of the last $m$ outputs of the network we apply the softplus function
\begin{equation*}
    \operatorname{Softplus}(t) = \log(1 + e^t)
\end{equation*} 
and let $\sigma(x)$ denote resulting output. The entries of $\sigma(x)$ are the standard deviations in the mixture of Gaussians. The softplus function ensures that the entries of $\sigma(x)$ are positive and prevents them from becoming too small, which could cause numerical instability. Together, $w(x)$ and $\sigma(x)$ yield an estimate of the conditional probability density function given by
\begin{equation*}
    \hat{p}(y|x) = \sum_{i=1 }^m\frac{w(x)_i}{\sigma(x)_i}\phi\Big(\frac{y-y_i}{\sigma(x)_i}\Big).
\end{equation*}
In the above we could also learn the centers of the network: this procedure is exactly a mixture density network \cite{bishop1994mixture}, which has not been tested in this study.
In \cite{ambrogioni2017kernel}, the negative log-likelihood is used for training the network. It is also possible to use the CRPS for training, as a closed form expression is known for the CRPS of a mixture of Gaussians (\cite{doi:10.1256/qj.05.235}): if $F$ is the cumulative distribution function of a mixture of $m$ Gaussians with weights $w_1,\ldots,w_m$, means $\mu_1,\ldots,\mu_m$, and variances $\sigma_1^2,\ldots\sigma_m^2$, then 
\begin{equation*}
\operatorname{CRPS}(F,y) = \sum_{i=1}^{m}w_i A(y-\mu_i,\sigma_i^2)-\frac{1}{2}\sum_{i=1}^ {m}\sum_{j=1}^{m}w_i w_j A(\mu_i-\mu_j,\sigma_i^ 2+\sigma_j^ 2),
\end{equation*}
where 
$$A(\mu,\sigma^2) = \mu\Big(2\Phi\Big(\frac{\mu}{\sigma}\Big)-1\Big)+2\sigma\phi\Big(\frac{\mu}{\sigma}\Big)$$
and $\Phi$ is the cumulative distribution function of a standard Gaussian. To our knowledge the CRPS has not been used before for kernel mixture networks or quantized softmax. It has, however, been applied with success to train mixture density networks in \cite{d2018photometric}} \simon{and \cite{rasp2018neural}.} 

\subsection{\sjoerd{Fitting a truncated normal distribution}}

\sjoerd{The third method considered in this study uses a neural network to learn the parameters of the normal distribution that has been truncated at zero. As was discussed in the introduction, this distribution has been successfully used for post-processing of \kiri{wind speed} forecasts. As the basis for the method we construct a neural network with two output neurons. Let $\mu(x)$ denote the first output and let $\sigma(x)$ be the result of applying the softplus function to the second output. The corresponding estimate of the conditional probability density function is then given by}

\begin{equation*}
\hat{p}(y|x) = \frac{\frac{1}{\sigma(x)}\phi\Big(\frac{y-\mu(x)}{\sigma(x)}\Big)}{1-\Phi\Big(-\frac{\mu(x)}{\sigma(x)}\Big)}, \qquad \text{if } y>0,
\end{equation*}

and $\hat{p}(y|x)=0$ else. We can again use the log-likelihood and CRPS for training. If $F$ denotes the cumulative distribution function of a normal distribution truncated at zero, then the CRPS is  given by  

\begin{equation*}
\operatorname{CRPS}(F,y) = \frac{\sigma}{p^2}\left[   s p ( 2\Phi(s) +p -2 ) + 2p \phi(s) -\frac{1}{\sqrt{\pi}}\Phi\Big(\frac{\mu\sqrt{2}}{\sigma}\Big)   \right],
\end{equation*}

where $p = \Phi(\frac{\mu}{\sigma})$ and $s = \frac{y-\mu}{\sigma}$ \citep{thorarinsdottir2010probabilistic}.

%Similar to what is described for the Kernel Density networks one wants the standard deviation to be positive and non-zero. A softmax activation function is therefore a good choice here as well. Furthermore one needs to be careful with the initialization of the model. An initial estimate for the standard deviation can, especially when using negative log-likelihood as loss function, lead to exploding gradients and as a result a diverging model. If the CRPS is used as a loss function, the initialization is less of an issue. This is due to the fact that for a point forecast the CRPS is equal to the mean absolute error, for which the derivative is constant. This results in a lot more stable behaviour during the training phase. 

\bibliographystyle{ametsoc2014}
\bibliography{CNN_paper_revised2.bib}

\begin{table}[t]
     \centering
     \resizebox{0.6\textwidth}{!}{\begin{minipage}{1.2\textwidth}
     \begin{tabular}{|c|c|c|c|c}
     \hline
Station Number & Longitude & Latitude & Name\\
\hline
 209&         4.518&       52.465&       IJMOND\\
 215&         4.437&       52.141&        VOORSCHOTEN\\
 225&         4.555&       52.463&       IJMUIDEN\\
 235&         4.781&       52.928&       DE KOOY\\
 240&         4.790&       52.318&       SCHIPHOL\\
 242&         4.921&       53.241&       VLIELAND\\
 248&         5.174&       52.634&       WIJDENES\\
 249&         4.979&       52.644&       BERKHOUT\\
 251&         5.346&       53.392&      HOORN (TERSCHELLING)\\
 258&         5.401&       52.649&        HOUTRIBDIJK\\
 260&         5.180&       52.100&        DE BILT\\
 267&         5.384&       52.898&        STAVOREN\\
 269&         5.520&       52.458&        LELYSTAD\\
 270&         5.752     &  53.224  &      LEEUWARDEN\\
 273&         5.888     &  52.703  &      MARKNESSE\\
 275&         5.873     &  52.056  &      DEELEN\\
 277&         6.200     &  53.413  &      LAUWERSOOG\\
 278&         6.259     &  52.435  &      HEINO\\
 279&         6.574     &  52.750  &      HOOGEVEEN\\
 280&         6.585     &  53.125  &      EELDE\\
 283&         6.657     &  52.069  &     HUPSEL\\
 285&         6.399     &  53.575  &       HUIBERTGAT\\
 286&         7.150     &  53.196  &      NIEUW BEERTA\\
 290&         6.891     &  52.274  &      TWENTHE\\
 308&         3.379     &  51.381  &       CADZAND\\
 310&         3.596     &  51.442  &      VLISSINGEN\\
 312&         3.622     &  51.768  &      OOSTERSCHELDE\\
 313&         3.242     &  51.505  &       VLAKTE V.D. RAAN\\
 315&         3.998    &   51.447  &       HANSWEERT\\
 316&         3.694    &   51.657  &       SCHAAR\\
 319&         3.861    &   51.226  &       WESTDORPE\\
 323&         3.884    &   51.527  &       WILHELMINADORP\\
 324&         4.006    &   51.596  &      STAVENISSE\\
 330&         4.122    &   51.992  &      HOEK VAN HOLLAND\\
 331&         4.193    &   51.480  &      THOLEN\\
 340&         4.342    &   51.449  &      WOENSDRECHT\\
 343&         4.313    &   51.893  &      R'DAM-GEULHAVEN\\
 344&         4.447    &   51.962  &      ROTTERDAM\\
 348&         4.926    &   51.970  &      CABAUW\\
 350&         4.936    &   51.566  &      GILZE-RIJEN\\
 356&         5.146    &   51.859  &      HERWIJNEN\\
 370&         5.377    &   51.451  &      EINDHOVEN\\
 375&         5.707    &   51.659  &      VOLKEL\\
 377&         5.763    &   51.198  &      ELL\\
380  &       5.762    &   50.906   &   MAASTRICHT\\
391  &       6.197    &   51.498    &    ARCEN\\
\hline
\end{tabular}
\end{minipage}}
\caption{Location of weather stations in the Netherlands.}
\label{tab:loc_weather}
\end{table}

\begin{table}[t]
    \centering
    \begin{tabular}{|c|c|c|}
    \hline
    \multirow{3}{*}{Model Selection} & Fold 1 & October - December 2015 and January - March 2016\\\cline{2-3}& Fold 2
    & October - December 2016 and January - March 2017\\\cline{2-3}& Fold 3
    & October - December 2017 and January - March 2015\\
    \hline
    \multirow{2}{*}{Test set}     & 
    \multicolumn{2}{c|}{November - December 2018, January-March 2019}\\& \multicolumn{2}{c|}{and October - November 2019}\\
     \hline
    \end{tabular}
    \caption{Definition of the different subsets used in cross-validation and testing.}
    \label{tab:folds_}
\end{table}{}

%\newpage
\begin{table}[t!]
\centering
\begin{tabular}{|c l|}
    \hline
   1.& \bf{Sine and cosine of the wind direction at a height of 10 m};\\
    2.& \bf{Wind speed at a height of 10 m};\\
    3.& \bf{Surface roughness};\\
    4.& \bf{Meridional/zonal wind components at 925 hPa};\\
    5.& \bf{Mean sea level pressure};\\
    6.& Total kinetic energy;\\
    7.& Humidity at surface level;\\
    8.& Geopotential height 500 hPa;\\
    9.& Temperature at surface level;\\
    10.& Meridional and zonal windcomponents at 850 hPa;\\
    11.& Day of the year;\\
    \hline
\end{tabular}
\caption{Predictors considered in our hyperparameter search. The predictors that gave the most \sjoerd{skillful} forecasts in cross-validation are indicated in bold.}
\label{bo: pred1}
\end{table}

%\begin{table}[t]
%\centering
%\begin{tabular}{|c l|}
%    \hline
%   1.& Sine and cosine of the wind direction at a height %of 10 m;\\
%    2.& Wind speed at a height of 10 m;\\
%    3.& Surface roughness;\\
%    4.& Meridional/zonal wind components at 925 hPa;\\
%    5.& Mean sea level pressure;\\
%    \hline
%\end{tabular}
%\caption{Predictors that gave the most \sjoerd{skillful} %forecasts in cross-validation}
%\label{bo: pred2}
%\end{table}

\begin{table}[t!]
\begin{tabular}{cc}
    \begin{minipage}{.45\linewidth}
    \centering
\begin{tabular}{|c|}
    \hline
    Fully connected layer\\
    \hline
    Relu\\
    \hline
    Batch Normalization\\
    \hline
    Dropout\\
      \hline
    \end{tabular}
   \captionof*{table}{Dense Block}
   \centering
      \begin{tabular}{|c|}
    \hline
    Convolution layer\\
    \hline
    Relu\\
    \hline
    BatchNormalization\\
    \hline
    MaxPooling2D\\
      \hline
    \end{tabular}
   \captionof*{table}{Convolution\sjoerdr{al} Block}
    \end{minipage} &

    \begin{minipage}{.45\linewidth}
    \begin{tabular}{|c|c|}
    \hline
    Convolutional Input& \\
	Convolutional Block& \\
	Convolutional Block& \\
	Convolutional Block& Dense Input\\
    \hline
    Dense Block & Dense Block\\
    \hline
    \multicolumn{2}{|c|}{Dense Block}\\
    \multicolumn{2}{|c|}{Output layer}\\
  \hline
    \end{tabular}
    \end{minipage} 
\end{tabular}
\captionof{table}{Final convolutional neural network architectures, where the size of the fully connected layers and convolutional layers are as given in Table \ref{tab:conv_hyper}. The output layer is a fully connected layer where the size and activation functions depend on the method used.}
\label{tab: conv_architecture}
\end{table}{}

\newpage
\begin{table}[t]
    \centering
    \begin{tabular}{|c|c|c|}
        \hline
        Hyperparameter & NN & NN\_LR \\
        \hline
        Number of layers & 2 & 3\\

        Layer size & 106 & 106\\
        Batch size & 256 & 256\\
        
        Learning rate & $3.47*10^{-3}$ & $1.57*10^{-3}$\\
        Dropout rate & 0.030 & 0.188 \\
        Loss function & log-likelihood& log-likelihood\\
        Decay parameter & $5.0 * 10^{6}$& $8.4 * 10^{4}$\\
        $\sigma^    2$ noise & 0 & 0.315\\
\hline
    \end{tabular}
    \caption{Hyperparameters for the selected models.}
    \label{tab:nn}
\end{table}{}

\begin{table}[t]
    \centering
    \begin{tabular}{|c|c|c|c|}
    \hline
    Hyperparameter     &  CNN\_LR\_N0 & CNN\_LR\_KMN & CNN\_LR\\
    \hline
        Input \sjoerdr{grid size} & 100x100 & 60x60 & 60x60\\
        Variables & 1,2,3,4,5 & 1,2,3,4,5 & 1,2,3,4,5\\
        Layer\_size & 60 & 80 & 80\\
        Size of convolutional layers & 16 & 16 & 16 \\
        Batch size & 128 & 128 & 128\\
        Learning rate & 0.0013 &   0.00053 & 0.0007283\\
        Loss function & CRPS & CRPS & log-likelihood\\
        Dropout rate & 0.1028 &0.072 &0.0888\\
        Decay parameter& 2.633e-06&4.098e-5& 4.10e-07\\
        Noise &0.315 &0.26218 &0.322\\
        Number \sjoerdr{of} kernels & n/a & 60 & n/a\\
        \hline
        
    \end{tabular}
    \caption{Hyperparameters of CNNs}
    \label{tab:conv_hyper}
\end{table}

\begin{table}[t]
\centering
\begin{tabular}{|l | c | c | c |}
\hline
Method & Fold 3 & Fold 1 & Fold 2 \\
\hline
NN & 0.824 & 0.898 & 0.914\\
NN\_LR & 0.828 & 0.865 & 0.889\\
QRF & 0.814&  0.861& 0.888 \\
QRF\_LR & 0.819 & 0.871 & 0.900  \\
CNN\_LR\_KMN & 0.794 & 0.830 & 0.861\\
CNN\_LR\_N0 & 0.772 & \bf{0.806} & 0.848\\
CNN\_LR & \bf{0.769} & 0.810 & \bf{0.839}\\
\hline
\end{tabular}
\caption{Continuous ranked probability score of different methods in cross-validation, with bold values indicating the best scores. \sjoerdr{%The fold numbers are as defined in 
Fold 1, Fold 2 and Fold 3 refer to the verification fold in cross-validation \mauricer{(Table~\ref{tab:folds_}})}.}
\label{table;results}
\end{table}

\begin{table}[t]
    %\centering
      \resizebox{0.8\textwidth}{!}{\begin{minipage}{1.2\textwidth}
    \begin{tabular}{|c||c|c|c||c|c|c||c|c|c||c|c|c|}
    \hline
    &  \multicolumn{3}{c||}{RMSE} &\multicolumn{3}{c||}{MAE}
&\multicolumn{3}{c|}{CRPS}&\multicolumn{3}{c|}{Log score}\\
&Fold3 & Fold1 & Fold2&Fold3 & Fold1 & Fold2&Fold3&Fold1&Fold2&Fold3&Fold1&Fold2\\
\hline
NN &2.457 &2.331 &2.391 &1.144 & 1.117& 1.124 & 0.820 $\pm 0.012$ & 0.799 $\pm 0.011$& 0.809$\pm 0.011$ & 4.42 & 4.36 & 4.37\\
NN\_LR &2.204 &2.126 &2.176 &1.113 &1.089& 1.100 & 0.793 $\pm 0.012$ & 0.779$\pm 0.011$ & 0.786$\pm 0.012$ &3.99 & 3.98 & 3.98\\
QRF &2.244 &2.220 &2.245 & 1.080 &1.076 &1.085 & 0.782$\pm 0.012$ & 0.776$\pm 0.012$ & 0.779$\pm 0.012$ &4.03 & 4.02& 4.02\\
QRF\_LR &2.157 &2.151 &2.154 &1.089& 1.090& 1.087 & 0.780$\pm 0.012$ & 0.781$\pm 0.012$ & 0.780$\pm 0.012$ &4.05& 4.04& 4.04\\
CNN\_LR\_KMN &1.968& 1.886 &1.922 &1.056& 1.046& 1.053 & 0.752$\pm 0.011$ & 0.744$\pm 0.011$ & 0.748$\pm 0.011$& 3.96& 3.97& 3.95\\
CNN\_LR\_N0 &\bf{1.818}& 1.861 &2.117 &\bf{1.012} &1.029 &1.088 & \bf{0.722$\pm 0.011$} & 0.732$\pm 0.011$ & 0.770$\pm 0.013$& \bf{3.90}& 3.92& 3.96\\
CNN\_LR &1.851& \bf{1.814} &\bf{1.889} &1.014& \bf{1.004}& \bf{1.032} & 0.724$\pm 0.011$ &
\bf{0.718$\pm 0.011$} & \bf{0.733$\pm 0.011$} & 3.91& \bf{3.91}& \bf{3.92}\\
%std_cnn_n0:  [0.010916283948188853, 0.010791402034513557, 0.013424085853493826]
%std_cnn_qsm:  [0.011315034782780425, 0.010572911619841408, 0.0111723442837439]
%std_cnn_kmn:  [0.01118704404604901, 0.010542588509726702, 0.011077165667868578]
%std_fcnn_error:  [0.011762836241033511, 0.01136427841677087, 0.011517697809978293]
%std_fcnn  [0.011922444904270406, 0.011401941827899, 0.0112030445034778]
%qrf:  [0.012123457277438824, 0.011890955587714459, 0.011940941836624789]
%qrf_error:  [0.01186233222570606, 0.01157455511454424, 0.011823564721238435]
\hline
    \end{tabular}
    \end{minipage}}
    \caption{Results on the independent test set. \simon{Fold1, Fold2 and Fold3 refer to the fold that is left out of the training data} \sjoerdr{(Table \ref{tab:folds_})}. The standard deviation in the CRPS was estimated by block bootstrapping 1000 times. }
    \label{tab:results}
\end{table}{}

\newpage
\begin{table}[t]
\centering
\begin{tabular}{|l | c | c | c | c |}
\hline
Method & RMSE & MAE & CRPS & Log score\\
\hline
Climatology & 2.676&2.033&1.445& $\infty$\\
Linear Regression & 2.399& 1.170 & - & -\\ 
QRF & 2.217&  1.077& 0.776 & 4.02\\
QRF\_LR & 2.124& 1.081& 0.774 &4.03\\
CNN\_LR\_KMN & 1.905 & 1.034 & 0.740 & 3.96\\
CNN\_LR\_N0 & 1.891 & 1.037 & 0.735 & \bf{3.91}\\
CNN\_LR & \bf{1.889} & \bf{1.033} & \bf{0.731} & 3.93\\
\hline
\end{tabular}
\caption{The root mean squared error, mean absolute error, continuous ranked probability score and log score of different methods for the independent test set, trained on the \sjoerdr{full} training data set (Table \ref{tab:folds_}).}
\label{table: results_fullset}
\end{table}

\end{document}